\documentclass[letterpaper, 10 pt, journal, twoside]{ieeetran}

\author{Lu Gan$^{1}$, Jessy W. Grizzle$^{1}$, Ryan M. Eustice$^{1}$, and Maani Ghaffari$^{1}$%
\thanks{Toyota Research Institute provided funds to support this work. Funding for J. Grizzle and M. Ghaffari was in part provided by NSF Award No. 2118818. This work was also supported by MIT Biomimetic Robotics Lab and NAVER LABS. NVIDIA Corporation provided hardware support for this work. Finally, we would like to thank Ray Zhang and Yukai Gong for valuable discussions, and Tzu-Yuan Lin for data collection.} %
\thanks{$^{1}$The authors are with Robotics Institute, University of Michigan, Ann Arbor, MI 48109, USA.
        {\tt\footnotesize \{ganlu, grizzle, eustice, maanigj\}@umich.edu}}%
}

\title{Energy-based Legged Robots Terrain Traversability Modeling via Deep Inverse Reinforcement Learning}

\usepackage[usenames,dvipsnames,svgnames,table,xcdraw]{xcolor}
\usepackage{amssymb} %
\usepackage{amsmath} %
\usepackage{graphicx} %
\usepackage{caption}
\usepackage{subcaption}
\usepackage[noadjust]{cite} %
\usepackage[numbers,sort&compress]{natbib}
\usepackage{algorithm, algorithmicx, algpseudocode}

\usepackage[colorlinks=true,pdfpagemode=UseNone,citecolor=black,linkcolor=black,urlcolor=BrickRed]{hyperref}
\usepackage{cleveref} %
\usepackage{booktabs}
\usepackage{balance}
\usepackage{multirow}

\newcommand{\rd}[1]{{\textcolor{Black}{#1}}}

\begin{document}

\maketitle

\begin{abstract}
This work reports on developing a deep inverse reinforcement learning method for legged robots terrain traversability modeling that incorporates both exteroceptive and proprioceptive sensory data. Existing works use robot-agnostic exteroceptive environmental features or handcrafted kinematic features; instead, we propose to also learn robot-specific inertial features from proprioceptive sensory data for reward approximation in a single deep neural network. 
Incorporating the inertial features can improve the model fidelity and provide a reward that depends on the robot's state during deployment.
We train the reward network using the Maximum Entropy Deep Inverse Reinforcement Learning (MEDIRL) algorithm and propose simultaneously minimizing a trajectory ranking loss to deal with the suboptimality of legged robot demonstrations.
The demonstrated trajectories are ranked by locomotion energy consumption, in order to learn an energy-aware reward function and a more energy-efficient policy than demonstration. We evaluate our method using a dataset collected by an MIT Mini-Cheetah robot and a Mini-Cheetah simulator. The code is publicly available at \mbox{{\footnotesize \href{https://github.com/ganlumomo/minicheetah-traversability-irl}{https://github.com/ganlumomo/minicheetah-traversability-irl}}.}
\end{abstract}


\section{Introduction}

\begin{figure}
    \centering
    \includegraphics[width=0.49\columnwidth, trim={0cm 7.5cm 0cm 7.8cm}, clip]{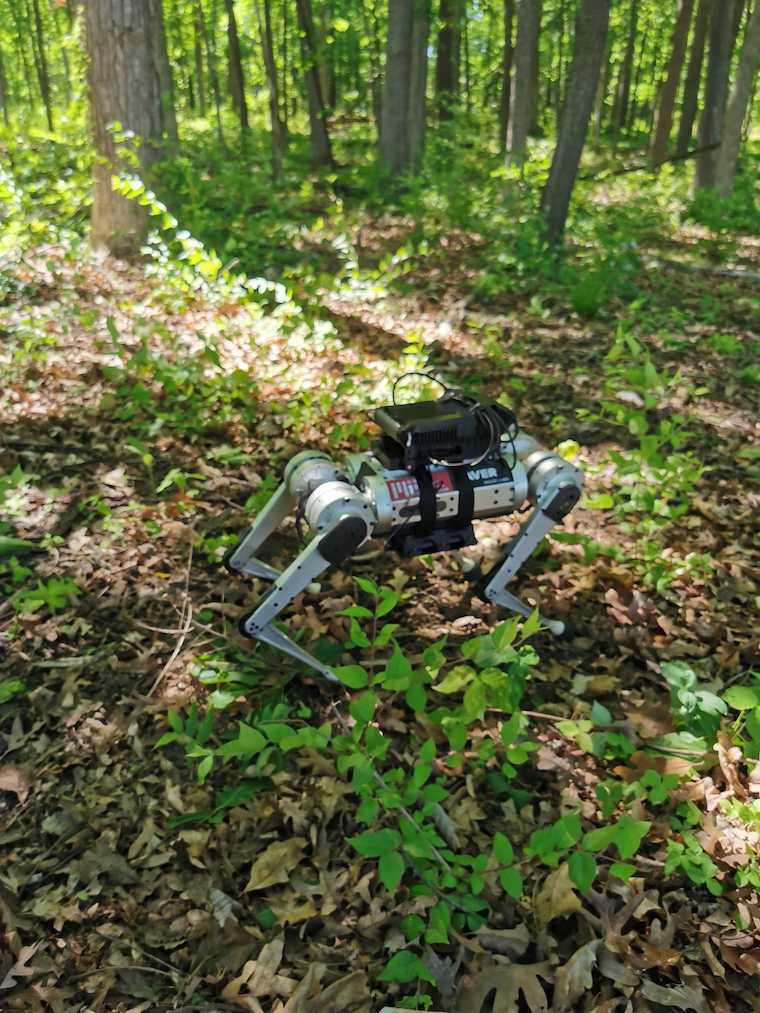}
    \includegraphics[width=0.48\columnwidth]{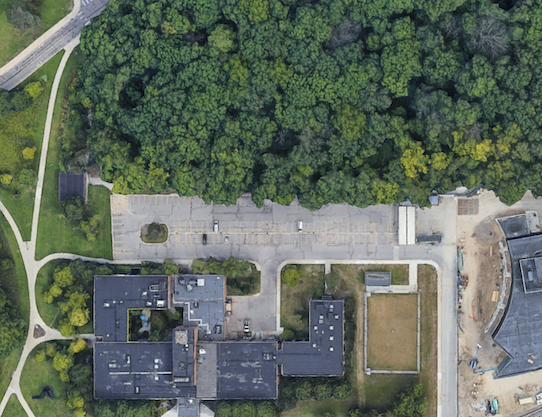}
    \caption{MIT Mini-Cheetah robot with a customized sensor suite and satellite map of the data collection environment for this work. The Mini-Cheetah robot is equipped with an Intel RealSense depth camera, an Inertial Measurement Unit (IMU), and an Nvidia Jetson Xavier computing device for data recording. Motion and task planning for legged robots in unstructured environments involve explicitly modeling the terrain traversability and solving a sequential decision-making problem to find the ``optimal'' behavior. For unknown environments, terrain traversability modeling using noisy measurements is itself a challenging problem due to the lack of \emph{a priori} knowledge of the surrounding terrain, the variability of terrain conditions, and the \emph{robot-specific} characteristic of traversability.}
    \label{fig:data-collection}
    \vspace{-4mm}
\end{figure}

\IEEEPARstart{F}{or} a robot to autonomously navigate an unknown and unstructured environment, end-to-end learning~\cite{nguyen2020autonomous, kahn2021badgr} and planning with terrain information~\cite{huang2021efficient, dashora2021hybrid} are two commonly applied approaches. Compared with end-to-end planning methods that directly map sensing to actions such as behavior cloning, planning with terrain information decomposes the complex problem, thus being more data-efficient and interpretable~\cite{guastella2021learning}.
The present work approaches legged robot autonomous exploration from the latter direction (see Fig.~\ref{fig:data-collection}).

A taxonomy of terrain traversability modeling methodologies includes proprioceptive approaches and exteroceptive approaches that further consist of appearance-based and geometry-based approaches~\cite{papadakis2013terrain}. Appearance-based approaches formulate terrain traversability modeling as a terrain type classification problem via image classification or semantic segmentation~\cite{gan2020bayesian}. However, such methods assume that each terrain type corresponds to the same level of traversability. Geometry-based approaches usually generate a geometric representation of the terrain from depth measurements and compute the traversability from local geometric features~\cite{shan2018bayesian}. Recent works~\cite{barnes2017find, gan2021multi} use appearance and geometry data to generate self-supervised traversability labels and train a neural network to infer the terrain traversability.

Traversability labels in exteroceptive-based methods ignore the robot's internal states while in motion across the terrain; hence, the degree of traversability is unknown, and the labels are usually binary (i.e., untraversable and traversable). In contrast, proprioceptive approaches can model terrain traversability as a continuous variable by defining a traversability cost using ground reaction score~\cite{wellhausen2019should}, vibration~\cite{bekhti2020regressed} or stability~\cite{faigl2019unsupervised} from a proprioceptive sensor modality. However, these metrics are chosen based on a user's domain knowledge and may not fully capture the traversability experienced by the robot itself.

Recently, Inverse Reinforcement Learning (IRL) for terrain traversability modeling has received increased interests, particularly for autonomous vehicles~\cite{wulfmeier2017large, zhang2018integrating, zhu2020off, jung2021incorporating}. Instead of using a predetermined metric to define traversability cost and associating it with terrain features, IRL-based methods aim to learn this cost directly from a robot's (traversing) behaviors. Doing so bypasses the effort and incompleteness of manually designing the cost function and parts of the learning pipeline. The resulting cost representations of IRL-based methods have also shown better robustness and scalability than the predetermined cost and the cost learned in supervised methods~\cite{wulfmeier2017large, osa2018algorithmic}.

Maximum Entropy Deep IRL (MEDIRL) exploits the representational capacity of neural networks for reward function approximation and has been successfully applied to traversability cost learning and trajectory forecasting for autonomous vehicles~\cite{wulfmeier2017large, zhang2018integrating, zhu2020off, deo2020trajectory, jung2021incorporating, wang2021inverse}. This work adopts the MEDIRL framework to tackle the problem of terrain traversability modeling for legged robots. A big challenge in IRL-based methods is the representation choice of the agent's state. A higher dimensional state generally leads to higher model fidelity but at the expense of an exponential increase in computational complexity. Therefore, most works incorporate higher-order robot states into the reward approximation network as extra features instead of expanding the dimension of the agent's state. However, encoding handcrafted robot kinematic features, such as estimated trajectory curvature~\cite{zhang2018integrating}, overlaid past trajectory~\cite{jung2021incorporating} and average velocities~\cite{deo2020trajectory}, has limited effect on improving the model fidelity due to the inherent approximation and inaccuracy in capturing the dynamics of the interactions with the environment.

Another challenge of applying the MEDIRL framework to legged robot applications is the suboptimality of expert demonstration. While maximum entropy methods are robust to some level of suboptimal demonstrations, they fundamentally seek a reward function that justifies the demonstrated behavior, leading to a policy that explains but rarely outperforms the demonstrator~\cite{brown2019extrapolating,arora2021survey}. Unlike a human driver, the demonstrator for legged robots typically lacks adequate feedback of locomotion performance to make the demonstration optimal or near-optimal. Further, legged robots are more likely to walk in unstructured environments where the optimal policy is difficult to find even for an expert. Thus, we propose to use robot proprioception as a source of feedback to rank each demonstrated trajectory, and introduce a \emph{trajectory ranking loss} in reward learning to learn a reward function that could extrapolate beyond suboptimal demonstrations and a policy outperforming the demonstrator.

This work 
offers the following major contributions:
\begin{enumerate}
    \item We propose a deep reward approximation network in the context of IRL that can learn robot inertial features from raw sensor measurements and incorporate the learned features into the reward map.
    \item We extend the MEDIRL framework into a Trajectory-ranked MEDIRL framework, and use locomotion energy as the trajectory preference label to learn an energy-aware reward map for legged robots.
    \item We conduct extensive experiments using real data collected by a quadruped robot in a campus environment, and evaluate our method using a robot simulator.
\end{enumerate}

\section{Related Work}
\label{sec:related_work}

This work is inspired by --- and bridges the gap between --- legged robots terrain traversability modeling and IRL-based traversability cost learning methods for autonomous vehicles.

\subsection{Terrain Traversability Modeling for Legged Robots}

Compared with ground vehicles, legged robots can traverse extreme, off-road, and unstructured terrains. Terrain traversability modeling is crucial for legged robots path planning and foot placement, however, remains challenging due to the wide variation in terrain conditions and robot models.

\citet{wellhausen2019should} propose a self-supervised learning framework to predict terrain properties for a quadruped robot. They compute a \emph{ground reaction score} from recorded force-torque measurements and associate the score with corresponding image data via foothold projection. The automatically annotated images are used to train a score regression network. Similarly, \citet{faigl2019unsupervised} also use robot proprioceptive features to characterize three traversal costs, i.e., mean instantaneous power consumption, mean forward velocity and attitude stability, and attach those costs to the corresponding exteroceptive features for regression.

\citet{bednarek2019touching} propose to use machine learning methods for tactile terrain classification. Later, a neural network is proposed to take variable-length signals such as forces and torques and output terrain classes for robot localization~\cite{buchanan2021navigating}.
\citet{fan2021step} probabilistically model the traversability cost as a combination of multiple risk factors determined by the terrain geometry and robot states.
Our method relates to these works as both exteroception and proprioception are exploited to model the terrain traversability for legged robots. But instead of manually defining the traversability cost function, we learn it from demonstration.

\subsection{IRL for Traversability Cost Learning}

Early works include applying the Maximum Margin Planning (MMP) framework to traversal cost modeling using overhead terrain data~\cite{silver2010learning}. Later, the MEDIRL framework~\cite{wulfmeier2015maximum, wulfmeier2017large} is proposed to exploit the power of deep neural networks in expressing nonlinear reward functions and becomes a common paradigm for traversability cost learning.

In~\cite{wulfmeier2017large}, the reward function is modeled using a fully convolutional network with the inputs of environmental features for each state, i.e., mean height, height variance, and a binary visible indicator. Vehicle kinematics is gradually considered in the MEDIRL framework in addition to environmental context to improve the model fidelity. \citet{zhang2018integrating} integrate the discretized vehicle's past velocity and trajectory curvature into the reward network. Similarly, the reward network in~\cite{deo2020trajectory} takes the encoded robot's motion features as inputs. \citet{jung2021incorporating} propose to incorporate inertial context into the traversability map by overlaying the past trajectory on an occupied grid map. \citet{zhu2020off} encode vehicle kinematic constraints into convolution kernels to accelerate the computation. We argue that handcrafted kinematic features are limited in improving the model fidelity; instead, our reward network learns the inertial features from raw sensor measurements.

\section{Methodology}
\label{sec:methodology}

\subsection{Problem Statement}

We model the process of a legged robot walking over local terrain as an agent following a Markov Decision Process (MDP). An MDP is defined as \mbox{$\mathcal{M} = \{\mathcal{S}, \mathcal{A}, \mathcal{T}, \gamma, r\}$} consisting of a set of states $\mathcal{S}$, actions $\mathcal{A}$, transition probabilities $\mathcal{T}$, a discount factor \mbox{$\gamma \in [0, 1)$}, and a reward function \mbox{$r: \mathcal{S} \rightarrow \mathbb{R}$}. A trajectory is defined as a sequence of state-action pairs \mbox{$\tau: = \{(s_0, a_0), (s_1, a_1), ..., s_{|\tau|}\}$} followed by the agent, and a demonstration is a set of trajectories under a policy \mbox{$\mathcal{D} = \{\tau_0, \tau_1, ..., \tau_{|\mathcal{D}|}\}$}. Given an MDP, a forward RL problem seeks the optimal policy \mbox{$\pi^\ast: \mathcal{S} \to \mathcal{A}$} that maximizes the expected discounted reward (return) \mbox{$J(\pi) = \mathbb{E}[\sum_{t=0}^{\infty} \gamma^t r_t | \pi]$}. We consider the IRL problem: Given an MDP$\backslash r$ and expert demonstration $\mathcal{D}$, to recover the underlying reward function that explains the demonstration, i.e., the reward function under which the demonstrated behavior is optimal. Specifically, we estimate the traversability cost a legged robot will get when traversing a location as the negative reward of that state using Maximum Entropy IRL (MEIRL).

However, the standard MEIRL formulation requires a pre-defined absorbing goal state (by holding the value of goal state $V(s_{goal})$ fixed to 0)~\cite{wulfmeier2015maximum, zhu2020off}, learning a reward function that is goal-conditioned. Other adapted methods relax the requirement~\cite{zhang2018integrating, jung2021incorporating}, but the demonstrated goal state is implicitly learned by the reward function through value iteration and policy propagation. This reward function is naturally suited to trajectory forecasting application, but less generalizable as a traversability cost map. To learn a goal-independent traversability cost map for exploration, we follow a reformulated MEIRL framework in~\cite{deo2020trajectory} which decouples the goal-conditioned reward map into a path reward map and a goal reward map. We define our traversability cost map as the negative path reward map, and it can also be used with a user-defined goal reward map to plan optimal paths to goal states other than the inferred ones.

Our MDP formulation is as follows.
\noindent \textbf{State~space}: We discretize the local terrain around a robot into a 2D grid of a certain resolution. Each cell in the grid could be a location the robot traverses (a path state), or the goal of the robot (a goal state). Thus, our state space is defined as $\mathcal{S}=\{\mathcal{S}_p, \mathcal{S}_g\}$, where $\mathcal{S}_p$ is the set of path states and $\mathcal{S}_g$ is the set of goal states. For instance, the state space of a $m \times m$ grid includes $m^2$ path states and $m^2$ goal states, and $2m^2$ states in total.
\noindent \textbf{Action space}: The actions of the agent are then chosen to be \mbox{$\mathcal{A} = \{up, down, left, right, end\}$} to allow the robot to transit from a path state to an adjacent path state, or to terminate at a goal state in that cell. It can also be easily extended to include 8 directions with evenly spaced 45-degree intervals.
\noindent \textbf{Transition function}: We define a \emph{deterministic} transition function \mbox{$\mathcal{T}:\mathcal{S}_p \times \mathcal{A} \rightarrow \mathcal{S}$} for the agent. The goal states do not have transitions as they are terminal.
\noindent \textbf{Rewards}: Two reward functions are defined, i.e., a path reward function \mbox{$r_p: \mathcal{S}_p \rightarrow \mathbb{R}$} as the reward the agent gets when it transits to a path state, and a goal reward function \mbox{$r_g: \mathcal{S}_g \rightarrow \mathbb{R}$} as the reward received when the agent terminates at a goal state. For instance, the total reward for a trajectory \mbox{$\tau = \{(s_{p0}, a_0),(s_{p1}, end), s_{g1}\}$} is \mbox{$r_{p0} + \gamma r_{p1} + \gamma^2 r_{g1}$}, where $\gamma$ is the defined discount factor.
\noindent \textbf{Traversability costs}: The traversability cost of each cell in the grid is defined as the negative path reward of the corresponding path state \mbox{$c=-r_p: \mathcal{S}_p \rightarrow \mathbb{R}$}.
The learned goal-independent traversability cost map can be used with a user-defined goal reward map (by setting high rewards for some goal states) for planning to a specific goal state or the goal reward map inferred by the reward network for exploration.

\subsection{Maximum Entropy Deep IRL}

MEIRL is formulated to address the challenge of reward ambiguity (i.e., multiple rewards can explain the same behavior) by treating trajectories with higher rewards as exponentially more likely~\cite{ziebart2008maximum}. Under the same assumption, MEDIRL~\cite{wulfmeier2015maximum} further approximates the reward function using a deep neural network parameterized by $\theta$, i.e, \mbox{$\hat{r}_{\theta}(s) = f(\phi(s);\theta)$}, where $\phi(s)$ denote the features of state $s$. It is then mathematically formulated as a maximum-a-posteriori (MAP) estimation problem that maximizes the logarithmic joint posterior distribution of the demonstration and network parameters under the predicted reward:
\begin{align}
\label{eq:medirl}
    \mathcal{L}(\theta) &= \log \mathbf{P}(\mathcal{D}, \theta | r_{\theta}) 
    = \log \mathbf{P} (\mathcal{D} | r_{\theta}) + \log \mathbf{P} (\theta) \nonumber \\
    &=: \mathcal{L}_\mathcal{D} + \mathcal{L}_{\theta}, 
\end{align}
which can be divided into a demonstration term $\mathcal{L}_\mathcal{D}$ and a model regularizer $\mathcal{L}_{\theta}$. Applying the chain rule, the reward network can be trained by backpropagating the following MEDIRL gradient with a regularization technique:
\begin{equation}
\label{eq:svf}
\frac{\partial \mathcal{L}_{\mathcal{D}}}{\partial \theta} = \sum_{\tau \in \mathcal{D}} (D_{\tau} - D_{r}) \frac{\partial r_{\theta}}{\partial \theta},
\end{equation}
where $D_{\tau}$ are the demonstrated State Visitation Frequencies (SVF) computed from the training data, and $D_{r}$ are the expected SVF of the policy given the predicted reward~\cite{ziebart2008maximum, wulfmeier2017large}. In this work, we use the Approximate Value Iteration and Policy Propagation algorithm in~\cite{deo2020trajectory} to compute $\mathcal{D}_r$ given the path and goal reward prediction, where $V(s_{goal})$ are set as the current predicted goal rewards instead of 0 at each iteration.

\begin{figure*}
    \centering
    \includegraphics[width=1.6\columnwidth, trim={0cm 2.0cm 0.1cm 0.7cm}, clip]{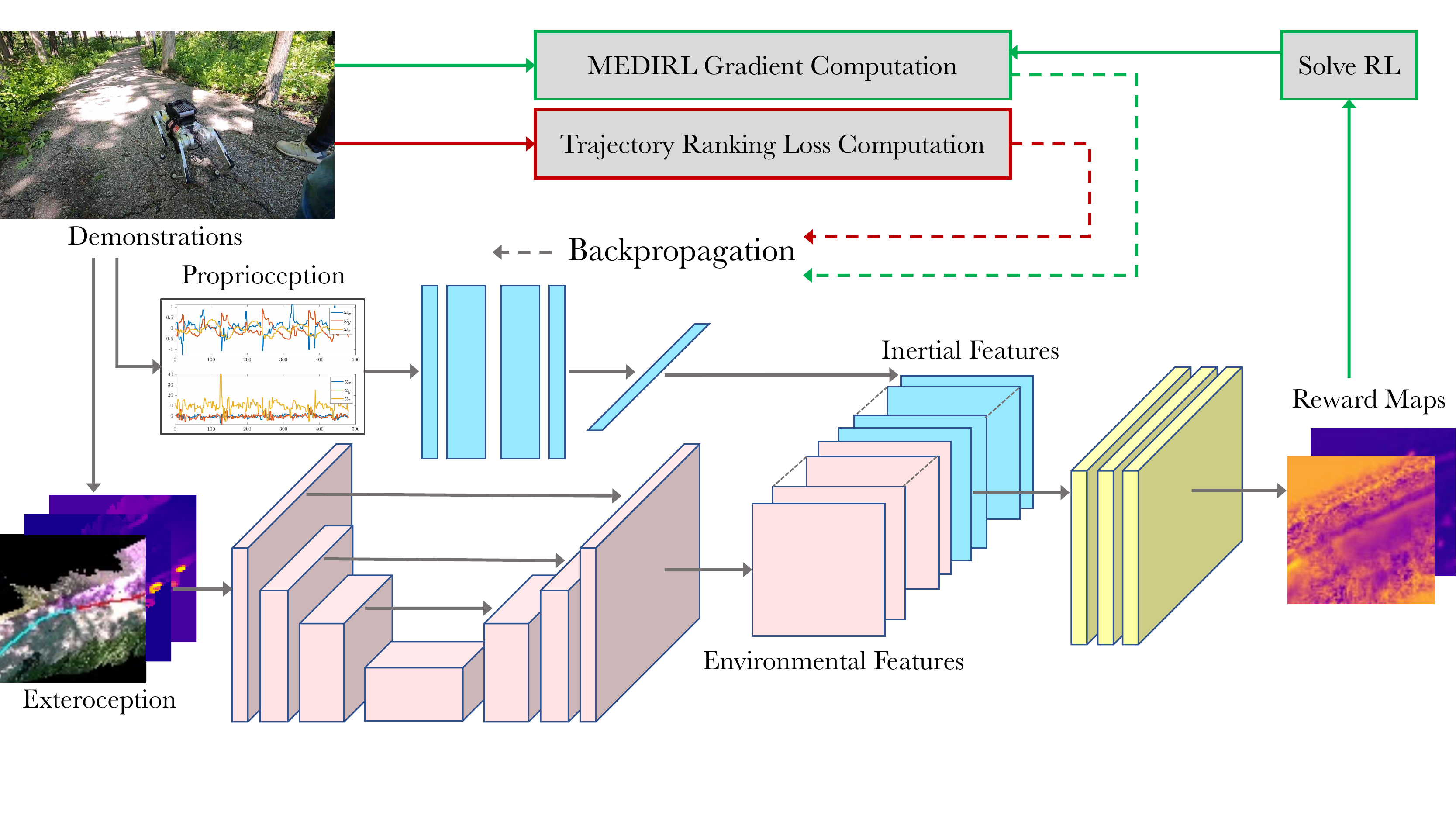}
    \caption{The proposed deep reward network architecture and training process for terrain traversability cost learning. The network has two stages: 1) feature learning and 2) feature fusion (yellow). To extract useful features from both proprioceptive and exteroceptive sensory data, the first stage consists of an inertial branch (blue) taking inputs of raw IMU signals and a ResUNet-based environmental branch (pink) taking inputs of the elevation map, elevation variance map and color map. The outputs of both branches are concatenated and fused in the second stage to approximate the reward maps. The network is trained in an end-to-end manner by backpropagating both MEDIRL gradient (green flow) and trajectory ranking loss (red flow). The average inference time for the reward network is 4.6e-04 $\sec$, and 0.018 $\sec$ for solving the RL using a GPU.}
    \label{fig:network-architecture}
\end{figure*}

\subsection{Learning Inertial Features: Network Architecture}

Existing works integrate environmental context and manually encoded robot kinematics into the reward network~\cite{zhang2018integrating, jung2021incorporating, deo2020trajectory}. In these works, the reward function can be written as \mbox{$\hat{r}_{\theta}(s) = f(\phi_e(s), \phi_k(s); \theta)$}, where $\phi_e(s)$ and $\phi_k(s)$ are the handcrafted environmental features and kinematic features of the state $s$. In this work, we propose a network architecture that can learn the robot inertial features directly from proprioceptive sensory data, and integrate them with environmental features in reward approximation: \mbox{$\hat{r}_{\theta}(s) = f(\phi_e(s), \phi_i(s; \theta); \theta)$}. Specifically, we learn a path reward $\hat{r}_{p\theta}(s)$ and a goal reward $\hat{r}_{g\theta}(s)$ using a network with shared parameters.

Inspired by the works~\cite{buchanan2021navigating, lin2021legged} where 1D convolutional modules are employed to efficiently process temporal proprioceptive sensory data of legged robots, we design an inertial branch for the reward network using similar structures. The branch consists of 2 convolutional blocks and 1 fully connected layer. Within the convolutional block, two 1D convolutional layers are followed by a 1D max pooling layer. 
Dropout layers are also used as the model regularizer $\mathcal{L}_{\theta}$ in~\eqref{eq:medirl}~\cite{wulfmeier2017large}. 
For simplicity, we choose fixed-length Inertial Measurement Unit (IMU) signals as the inputs to this branch. 

To model traversability, the geometry and appearance of the local terrain also contain critical information. Thus, another branch of the network aims to incorporate environmental context into the reward learning. Similar to previous works~\cite{wulfmeier2017large, zhang2018integrating, zhu2020off, deo2020trajectory}, we choose the elevation map, elevation variance map and a bird's eye view color map as the environmental features. Due to the spatial nature of environmental inputs, we adopt a ResUNet~\cite{zhang2018road} architecture for image segmentation as the environmental branch based on its quantitative and qualitative performance in experiments.

The proposed reward network therefore has a two-stage architecture, as shown in Fig.~\ref{fig:network-architecture}. The first stage comprises the aforementioned two branches to process inertial (proprioceptive) data and environmental (exteroceptive) data individually. The exteroceptive and proprioceptive data are synchronized using the recorded time stamps. For the environmental maps generated at time $t$ (when the robot is at the center of the maps), the associated IMU signals are collected from $t-\Delta t$ to $t$, where $\Delta t$ is a fixed time window. The inertial features extracted from the first branch are then upsampled and concatenated with the outputs of the environmental branch. Similar to~\cite{zhang2018integrating}, we encode the $n$-dimensional 1D output from the inertial branch as $n$-channel 2D feature maps. Additional 2-channel position encoding feature maps are also concatenated to break translation-invariance of the following convolutional filters. We refer readers to~\cite{zhang2018integrating} for more details. The second stage of the reward network is a block of 2D convolutions that fuses the two types of features into the final reward maps.

\subsection{Trajectory-ranked MEDIRL}

In the MEDIRL framework, the reward network is trained using stochastic gradient descent where the gradient comes from the differences between the demonstrated SVF and expected SVF in~\eqref{eq:svf}. Therefore, the reward function obtained by minimizing the differences will be the one best explaining the demonstration, and the corresponding policy will be suboptimal if the demonstration is suboptimal. To allow reward extrapolation when additional preference information is given, we propose to add a trajectory ranking loss to the MEDIRL framework and extend it to a Trajectory-ranked MEDIRL (T-MEDIRL) framework.

The trajectory ranking loss is inspired by the preference-based IRL methods~\cite{brown2019extrapolating,brown2020better} which seek a reward function that explains the \emph{ranking} over demonstrations (intention), in contrast to the \emph{demonstrations} (behavior), thereby allowing for reward generalization and extrapolation. Specifically, given a sequence of ranked trajectories, the objective is to learn a reward function that assigns higher returns to higher-ranked trajectories. It can thus be formulated as a classification problem that predicts whether a trajectory has a higher rank than another based on the predicted return (originally the predicted reward). It is worth mentioning that the intention is not to compare the optimality of two specific trajectories, but as a type of preference learning to regulate the reward learning process. The classifier can be trained by using the trajectory-rank pairs as data-label pairs and minimizing the pairwise trajectory ranking loss~\cite{brown2020better}:
\begin{equation}
\label{eq:trajectory-ranking-loss}
    \mathcal{L}_{i, j} = -\sum\limits_{\tau_i \prec \tau_j} \log \frac{\exp\sum\limits_{s \in \tau_j}r_{\theta, j}(s)}{\exp\sum\limits_{s \in \tau_i}r_{\theta, i}(s)+\exp\sum\limits_{s \in \tau_j}r_{\theta, j}(s)},
\end{equation}
where $r_{\theta, i}$ and $r_{\theta, j}$ are the predicted reward for trajectory $i$ and $j$, respectively, and \mbox{$\tau_i \prec \tau_j$} 
means that trajectory $j$ has a higher rank compared with trajectory $i$. In this work, $r_\theta$ in~\eqref{eq:trajectory-ranking-loss} is the path reward as the goal reward is only received once at the end of each trajectory, but we use $r_\theta$ for generalization purpose.

\begin{algorithm}[t]
\scriptsize 
\caption{{\small T-MEDIRL: Trajectory-ranked Maximum Entropy Deep Inverse Reinforcement Learning}}
\label{al:t-meirl}
\begin{algorithmic}[1]
\State \textbf{Input:} $\mathcal{S}, \mathcal{A}, \mathcal{T}, \gamma, \mathcal{D}, \alpha$
\State \textbf{Output:}
network parameters $\theta$

\Function{\textsc{MEDIRL-Grad}}{$r_{\theta}, \tau, \mathcal{S}, \mathcal{A}, \mathcal{T}, \gamma$}
    \State $\pi \gets \textsc{Approx.-Value-Iteration}(r_{\theta}, \mathcal{S}, \mathcal{A}, \mathcal{T}, \gamma)$
    \State $D_{\tau} \gets \textsc{Demo.-State-Visitations}(\tau, \mathcal{S})$
    \State $D_{r} \gets \textsc{Policy-Propagation}(\pi, \mathcal{S}, \mathcal{A}, \mathcal{T})$
    \State $\frac{\partial \mathcal{L}}{\partial r_{\theta}} \gets D_{\tau} - D_{r}$
    \State \Return $\frac{\partial \mathcal{L}}{\partial r_{\theta}}$
\EndFunction

\State
\State Randomly initialize $\theta^1$
\For{$n$ = 1 to $N$}
    \State Randomly choose $\tau_i$ and $\tau_j$ from $\mathcal{D}$
    \State $r_{\theta^n, i}(s) \gets f(\phi_i(s); \theta^n)$ for $\forall s \in \mathcal{S}$
    \State $r_{\theta^n, j}(s) \gets f(\phi_j(s);\theta^n)$ for $\forall s \in \mathcal{S}$
    \State \Comment{Forward reward network}
    \State $\frac{\partial \mathcal{L}_i}{\partial r_{\theta^n, i}} \gets \textsc{MEDIRL-Grad}(r_{\theta^n, i}, \tau_i, \mathcal{S}, \mathcal{A}, \mathcal{T}, \gamma)$
    \State $\frac{\partial \mathcal{L}_j}{\partial r_{\theta^n, j}} \gets \textsc{MEDIRL-Grad}(r_{\theta^n, j}, \tau_j, \mathcal{S}, \mathcal{A}, \mathcal{T}, \gamma)$
    \State \Comment{Compute MEDIRL gradient}
    \State $\mathcal{L}_{i, j} \gets -\sum\limits_{\tau_i \prec \tau_j} \log \frac{\exp\sum\limits_{s \in \tau_j}r_{\theta^n, j}(s)}{\exp\sum\limits_{s \in \tau_i}r_{\theta^n, i}(s)+\exp\sum\limits_{s \in \tau_j}r_{\theta^n, j}(s)}$
    \State \Comment{Compute trajectory ranking loss}
    \State Compute $\nabla_{\theta^n}\mathcal{L}_i, \nabla_{\theta^n}\mathcal{L}_j, \nabla_{\theta^n}\mathcal{L}_{i,j}$ via chain rule
    \State $\theta^{n+1} \gets \theta^n + \alpha \nabla_{\theta^n}\mathcal{L}_i + \alpha \nabla_{\theta^n}\mathcal{L}_j - \alpha \nabla_{\theta^n}\mathcal{L}_{i,j}$
    \State \Comment{Update network parameters}
\EndFor
\end{algorithmic}
\end{algorithm}

\begin{figure*}[t]
    \centering
    \includegraphics[width=0.25\columnwidth, trim={6cm 9.6cm 6cm 9.5cm}, clip]{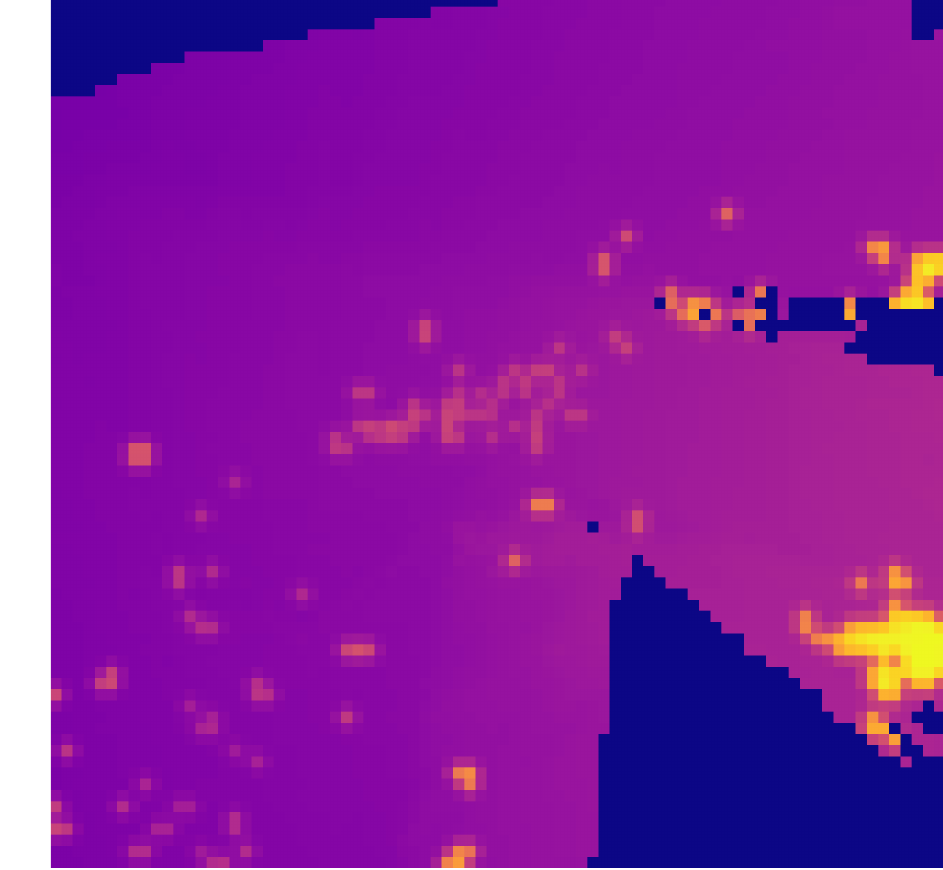}
    \includegraphics[width=0.25\columnwidth, trim={6cm 9.6cm 6cm 9.5cm}, clip]{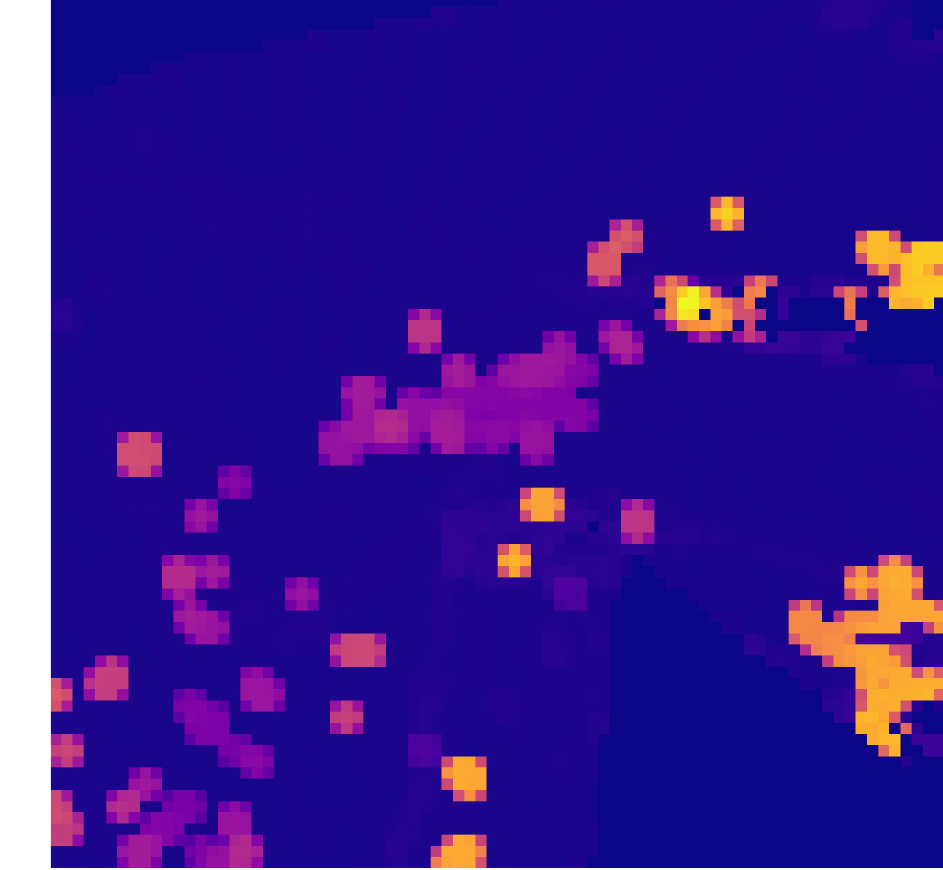}
    \includegraphics[width=0.25\columnwidth, trim={6cm 9.6cm 6cm 9.5cm}, clip]{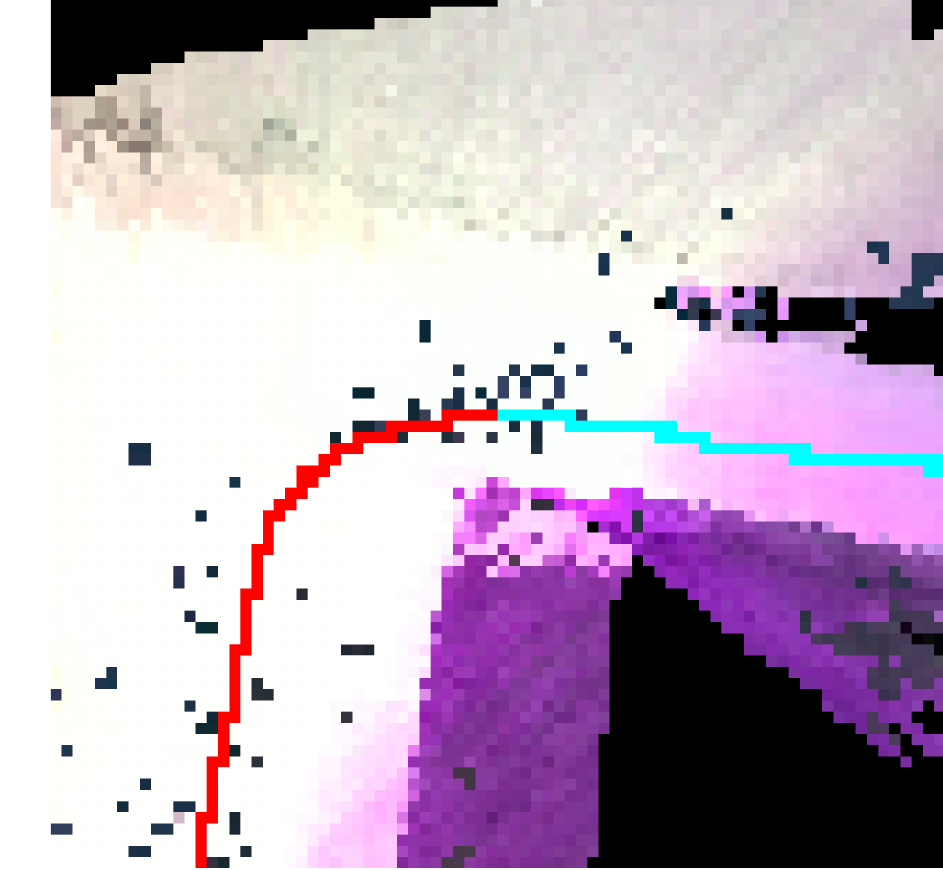}
    \includegraphics[width=0.41\columnwidth, trim={2cm 0cm 0cm 0cm}, clip]{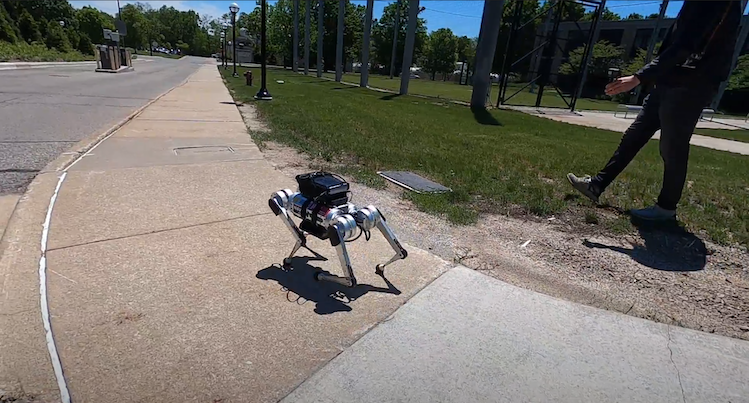}
    \includegraphics[width=0.36\columnwidth, trim={2cm 8.47cm 2cm 8cm}, clip]{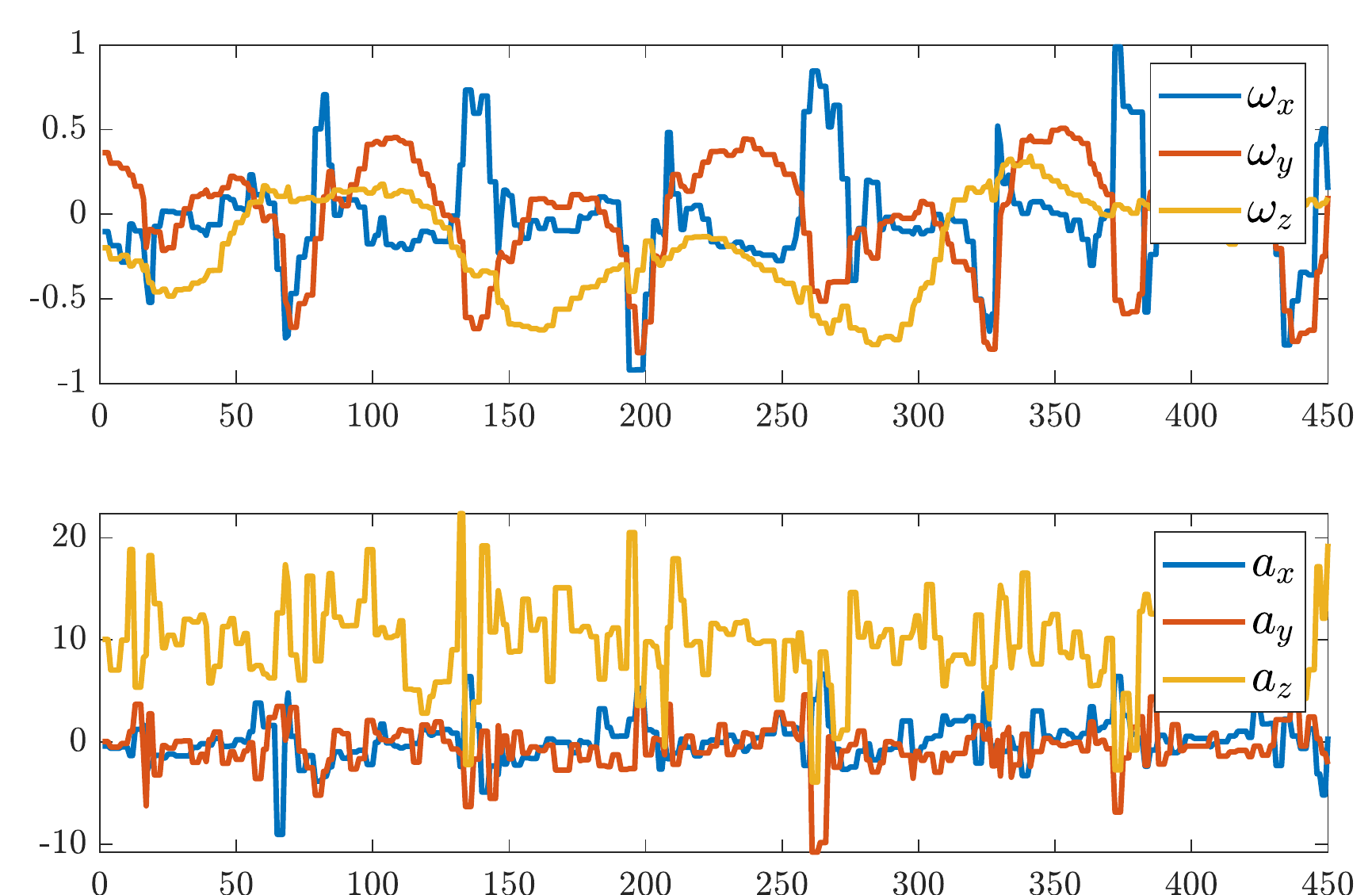} \\
    \includegraphics[width=0.25\columnwidth, trim={6cm 9.6cm 6cm 9.5cm}, clip]{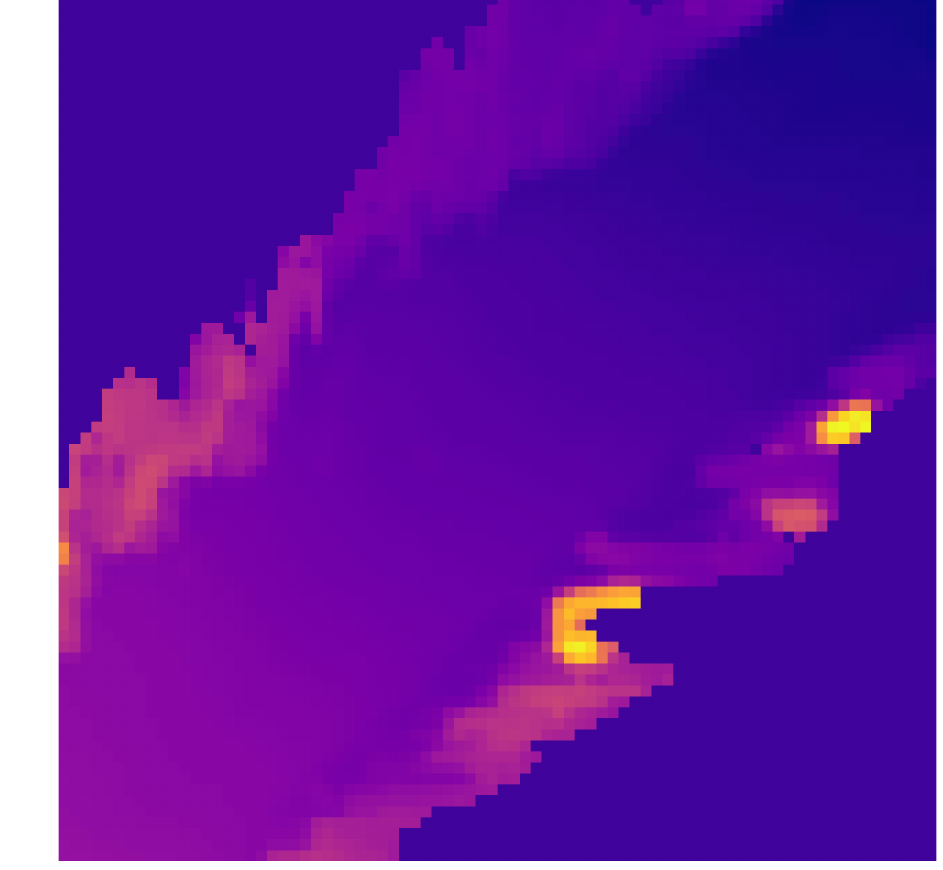}
    \includegraphics[width=0.25\columnwidth, trim={6cm 9.6cm 6cm 9.5cm}, clip]{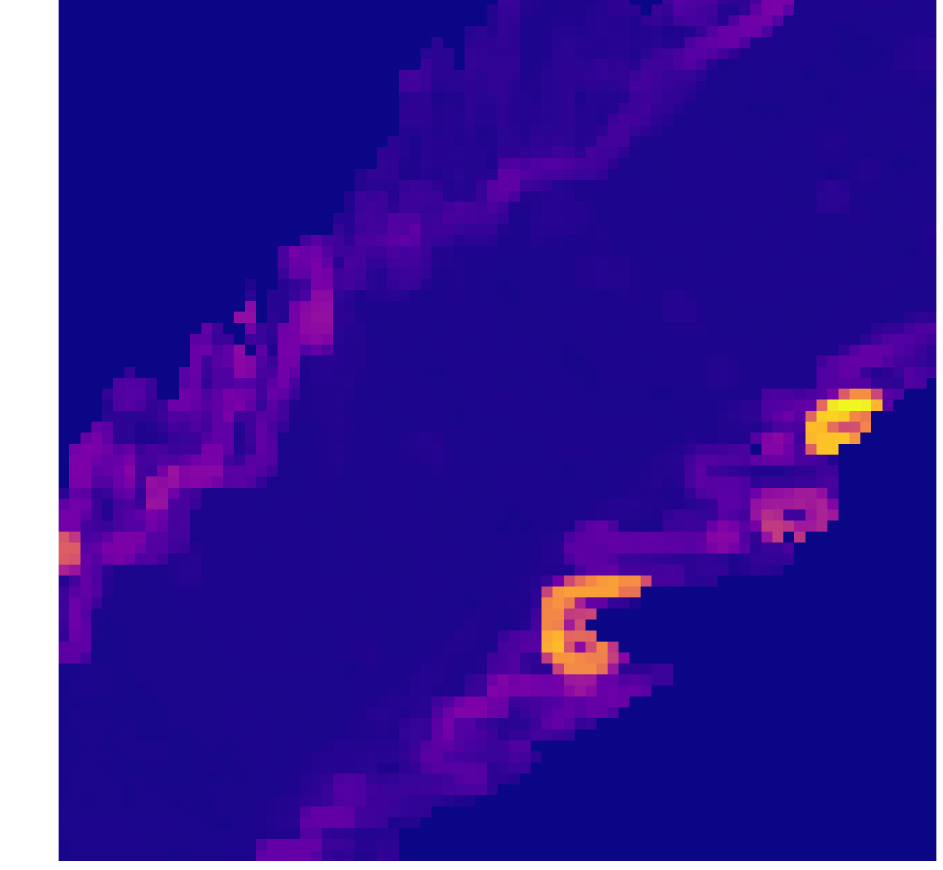}
    \includegraphics[width=0.25\columnwidth, trim={6cm 9.6cm 6cm 9.5cm}, clip]{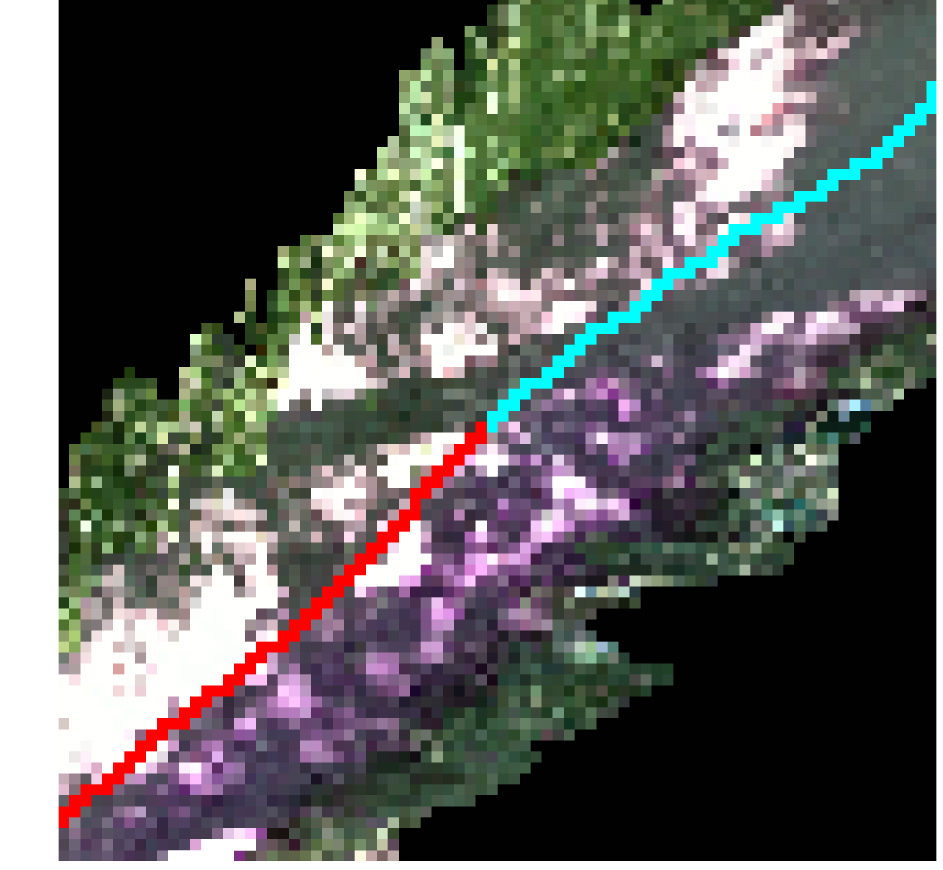}
    \includegraphics[width=0.41\columnwidth, trim={2cm 0cm 0cm 0cm}, clip]{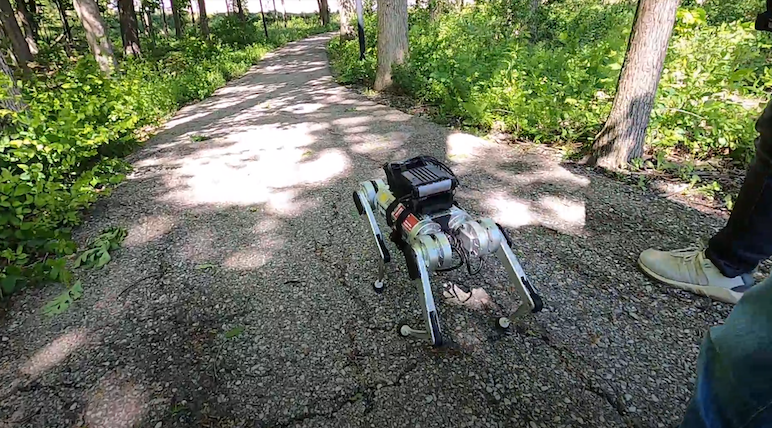}
    \includegraphics[width=0.36\columnwidth, trim={2cm 8.47cm 2cm 8cm}, clip]{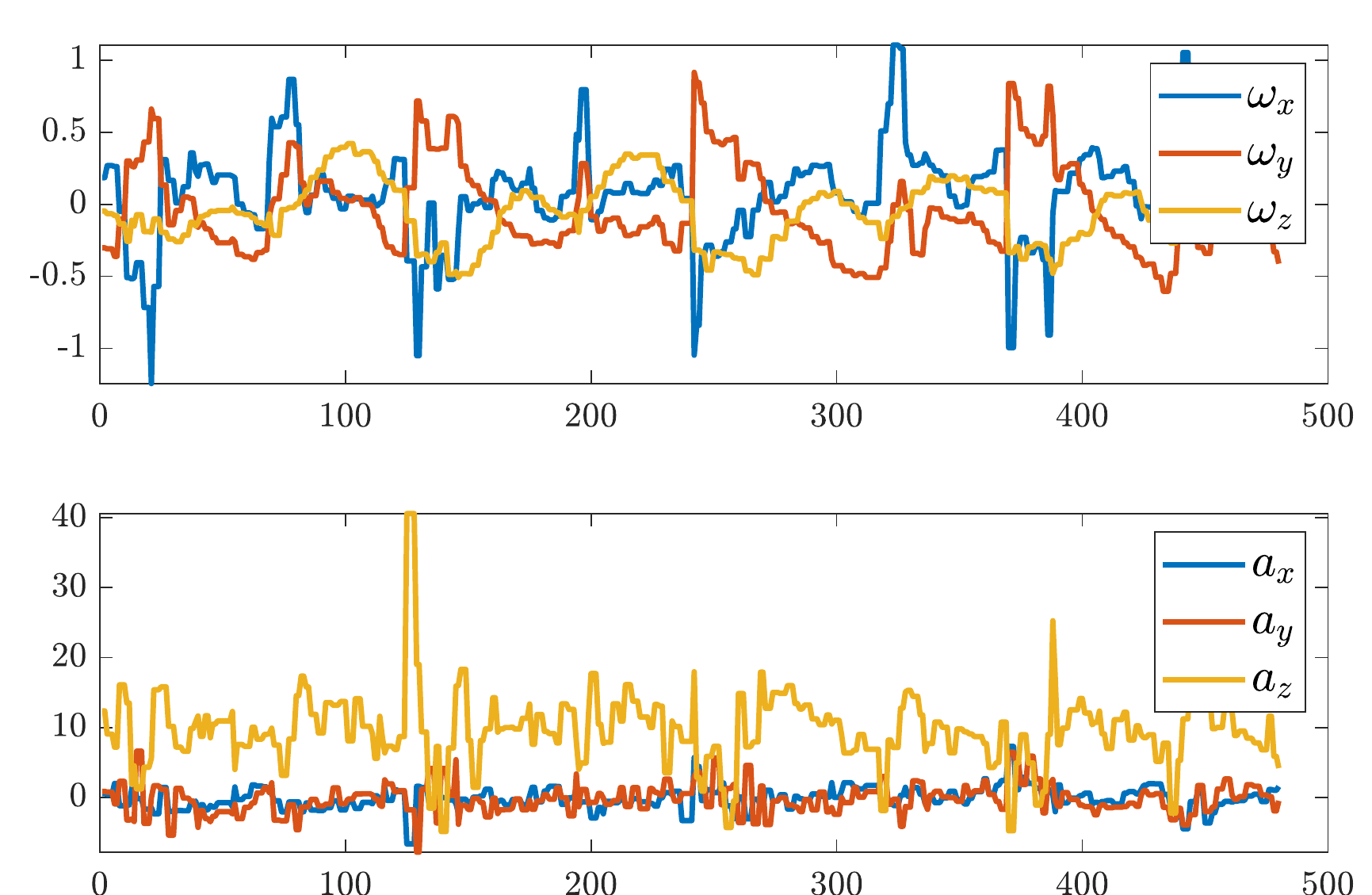} \\
    \begin{subfigure}{0.25\columnwidth}
        \includegraphics[width=1.0\columnwidth, trim={6cm 9.6cm 6cm 9.5cm}, clip]{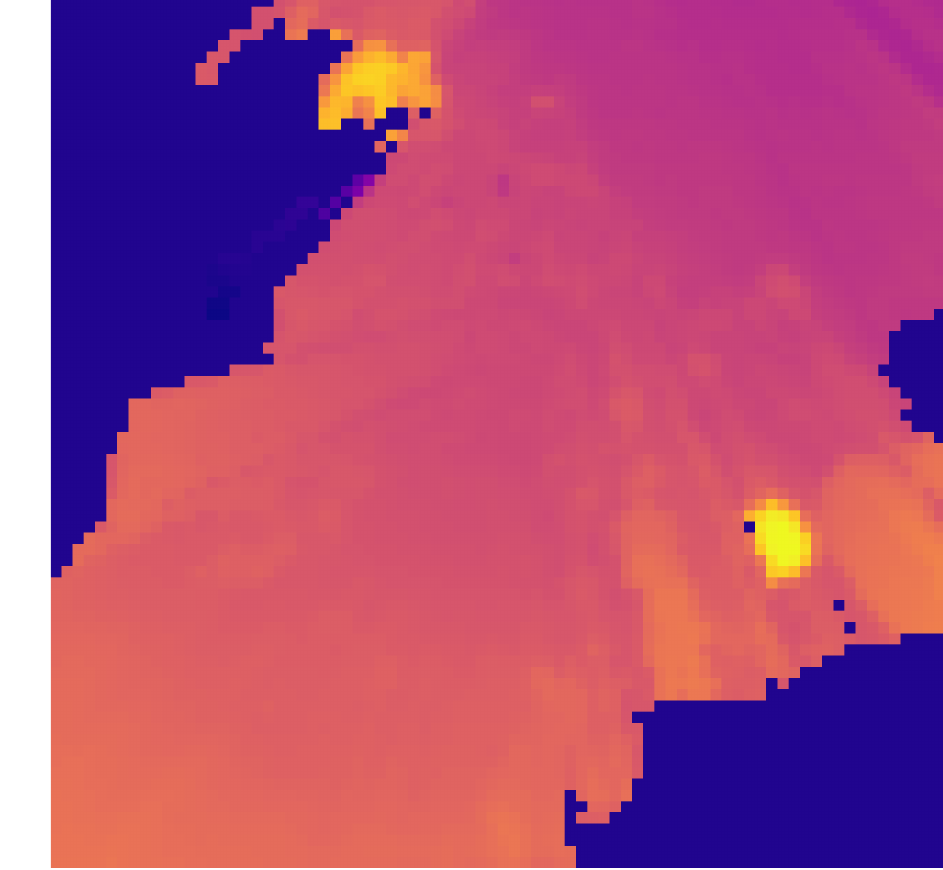}
        \caption{}
    \end{subfigure}
    \begin{subfigure}{0.25\columnwidth}
        \includegraphics[width=1.0\columnwidth, trim={6cm 9.6cm 6cm 9.5cm}, clip]{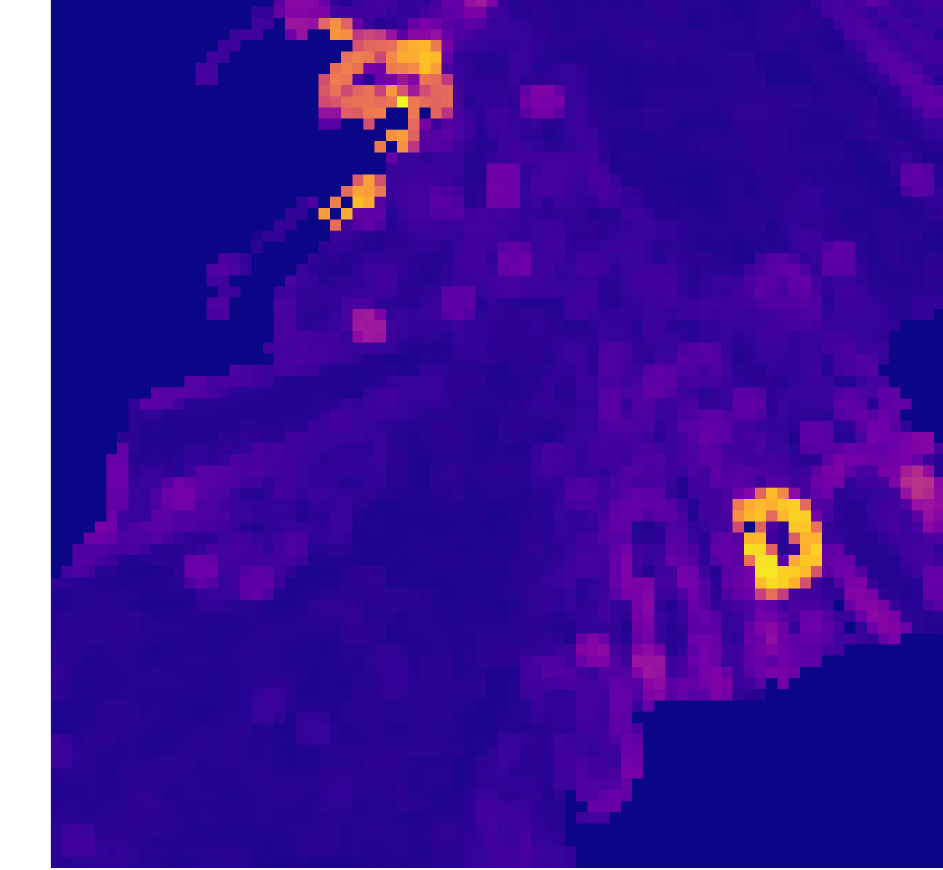}
        \caption{}
    \end{subfigure}
    \begin{subfigure}{0.25\columnwidth}
        \includegraphics[width=1.0\columnwidth, trim={6cm 9.6cm 6cm 9.5cm}, clip]{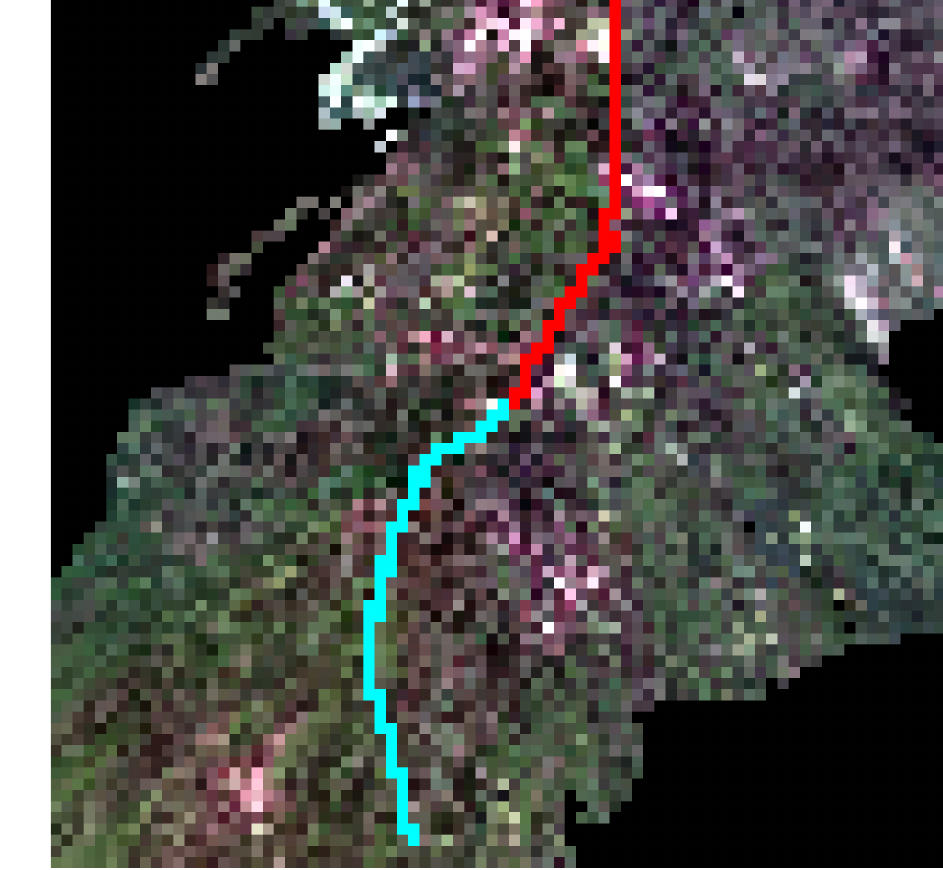}
        \caption{}
    \end{subfigure}
    \begin{subfigure}{0.41\columnwidth}
        \includegraphics[width=1.0\columnwidth, trim={2cm 0cm 0cm 0cm}, clip]{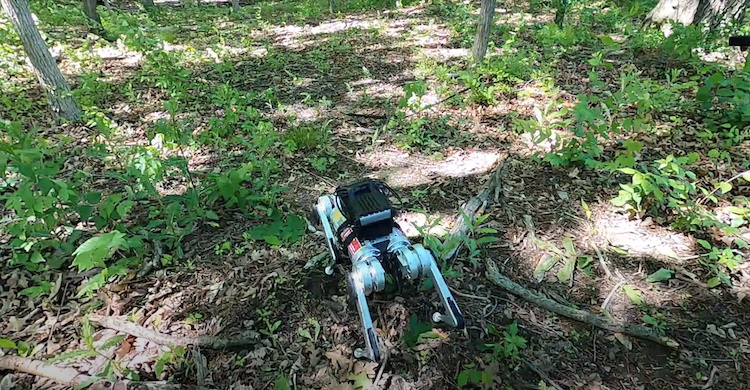}
        \caption{}
    \end{subfigure}
    \begin{subfigure}{0.36\columnwidth}
        \includegraphics[width=1.0\columnwidth, trim={2cm 8.47cm 2cm 8cm}, clip]{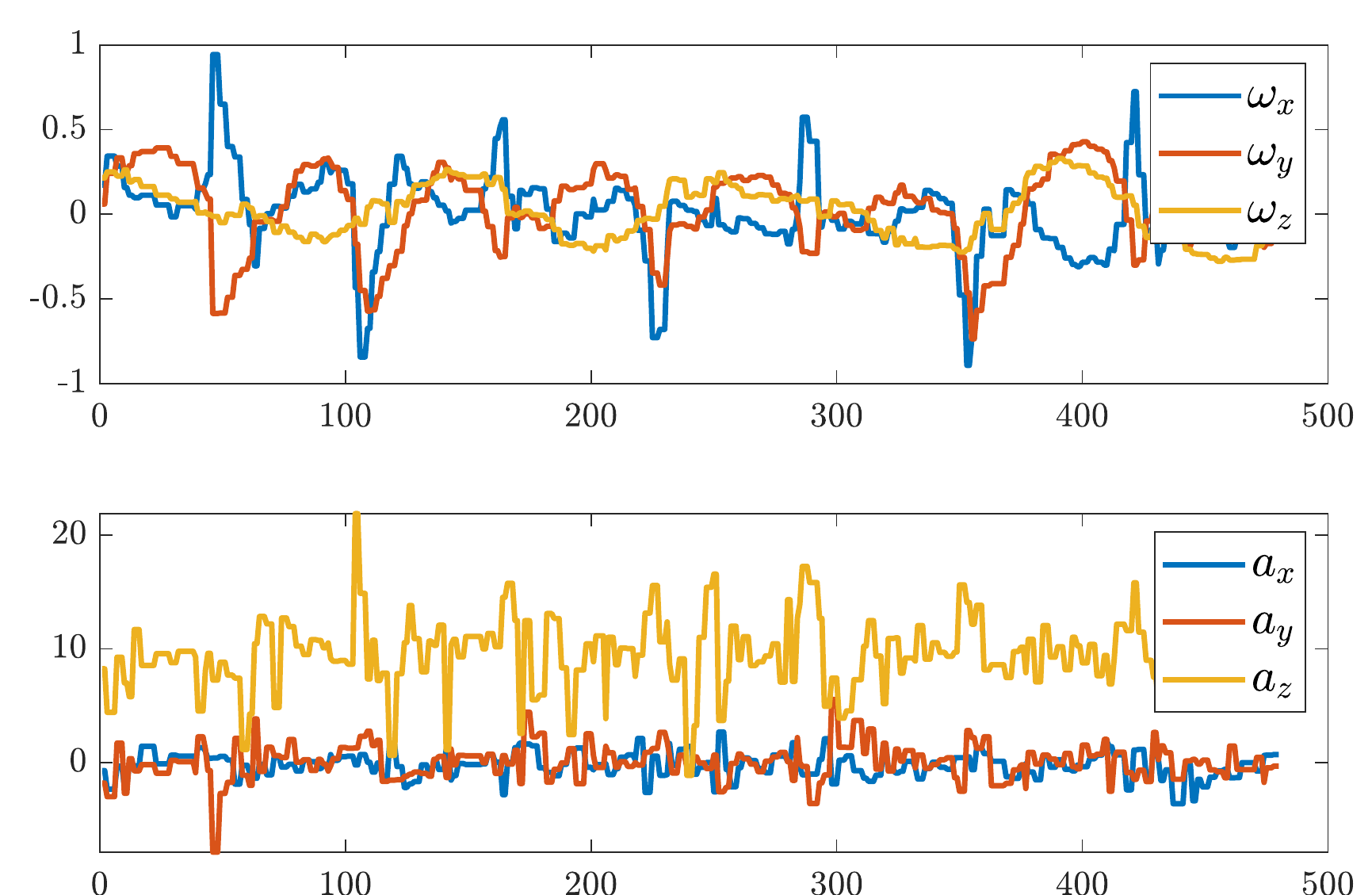}
        \caption{}
    \end{subfigure}
    \caption{Visualization of the generated data (network inputs) in different environments, for instance, sidewalk, trail and forest. (a) Elevation map. (b) Elevation variance map. (c) Color map. The demonstrated past (red) and future (cyan) trajectories are overlaid for visualization purpose and are not part of the feature map. (d) Third-person view of the current state (not an input). (e) The IMU signals collected within the past 0.5 $\sec$.}
    \label{fig:network-input}
\end{figure*}

We use the trajectory ranking loss to regulate the reward learning in the proposed T-MEDIRL algorithm (Algorithm~\ref{al:t-meirl}). Given the MDP$\backslash r$ and demonstration with ranked trajectories, the T-MEDIRL framework learns to approximate the reward function $r_{\theta}$ by backpropagating the MEDIRL gradient and the trajectory ranking loss together, as shown in Fig.~\ref{fig:network-architecture}. At each iteration, we randomly choose two (batches of) trajectories and obtain the corresponding predicted returns (reward functions) by forwarding the reward network. The per-trajectory MEDIRL gradient can be computed using the MEDIRL-Grad function, while the trajectory ranking loss is computed using~\eqref{eq:trajectory-ranking-loss} on a cross-trajectory basis. Finally, the network parameters are updated according to the learning rate $\alpha$ and gradients computed using the chain rule.

\subsection{Locomotion Energy Ranked Reward Extrapolation}

Energy cost is one of the major factors that can quantify the locomotion performance of legged robots, and an important consideration in path/motion planning. As a source of robot-terrain interaction, it also contains critical information about the traversed terrain. \citet{faigl2019unsupervised} use mean power consumption as an indicator of the terrain traversability, and \citet{walas2016terrain} optimize the gait parameters for a humanoid robot based on the energy expenditure and locomotion stability. However, the energy cost is usually unknown or unpredictable to the human demonstrator when collecting demonstrations of legged robot walking, leading to a severe suboptimality issue, i.e., the demonstrated trajectory might be feasible but energy inefficient.

To extrapolate beyond the suboptimal demonstrations for legged robots and learn a reward function that can lead to more energy-efficient path planning, we propose to use the Average Energy Consumption (AEC) of each trajectory for trajectory ranking in the T-MEDIRL framework. The AEC is computed by dividing the overall energy consumption of each trajectory by the trajectory length, where the overall energy consumption can be obtained using the recorded robot joint states as in~\cite{walas2016terrain}:
\begin{equation}
    e_{\tau} = \sum_{i=1}^n \langle \lvert \boldsymbol{u}_i \rvert, \lvert \Delta \boldsymbol{q}_i\rvert \rangle,
\label{eq:overall-energy}
\end{equation}
where $\lvert \cdot \rvert$ takes the element-wise absolute value, $\langle \cdot, \cdot \rangle$ denotes the dot product, $\boldsymbol{u} \in \mathbb{R}^{m}$ and $\Delta \boldsymbol{q} \in \mathbb{R}^{m}$ are the joint torque and joint displacement between two time stamps for $m$ joints, respectively, and $n$ is the total length of the signals collected along the entire trajectory $\tau$. 
In this work, if $\tau_j$ has a lower AEC value than $\tau_i$, then $\tau_i \prec \tau_j$.

\section{Experiments}
\label{sec:experiments}

\begin{figure*}[t]
    \centering
    \includegraphics[width=0.15\columnwidth]{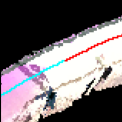}
    \includegraphics[width=0.15\columnwidth]{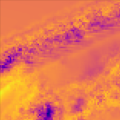}
    \includegraphics[width=0.15\columnwidth]{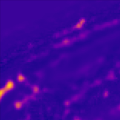}
    \includegraphics[width=0.15\columnwidth]{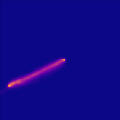}
    \includegraphics[width=0.15\columnwidth]{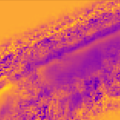}
    \includegraphics[width=0.15\columnwidth]{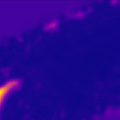}
    \includegraphics[width=0.15\columnwidth]{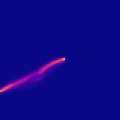}
    \includegraphics[width=0.15\columnwidth]{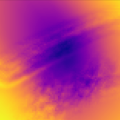}
    \includegraphics[width=0.15\columnwidth]{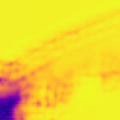}
    \includegraphics[width=0.15\columnwidth]{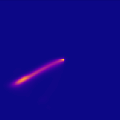}
    \includegraphics[width=0.01\columnwidth, height=0.14\columnwidth]{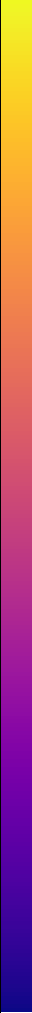}
    \\
    
    \includegraphics[width=0.15\columnwidth]{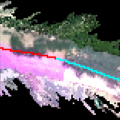}
    \includegraphics[width=0.15\columnwidth]{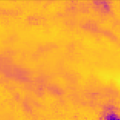}
    \includegraphics[width=0.15\columnwidth]{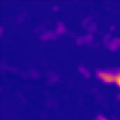}
    \includegraphics[width=0.15\columnwidth]{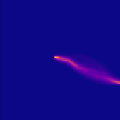}
    \includegraphics[width=0.15\columnwidth]{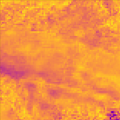}
    \includegraphics[width=0.15\columnwidth]{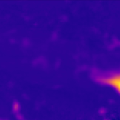}
    \includegraphics[width=0.15\columnwidth]{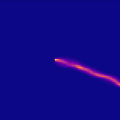}
    \includegraphics[width=0.15\columnwidth]{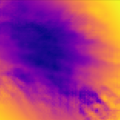}
    \includegraphics[width=0.15\columnwidth]{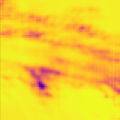}
    \includegraphics[width=0.15\columnwidth]{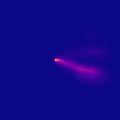}
    \includegraphics[width=0.01\columnwidth, height=0.14\columnwidth]{figs/revision/colorbar.png}
\\
    \includegraphics[width=0.15\columnwidth]{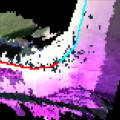}
    \includegraphics[width=0.15\columnwidth]{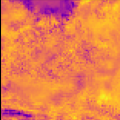}
    \includegraphics[width=0.15\columnwidth]{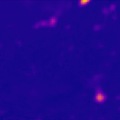}
    \includegraphics[width=0.15\columnwidth]{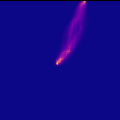}
    \includegraphics[width=0.15\columnwidth]{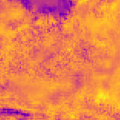}
    \includegraphics[width=0.15\columnwidth]{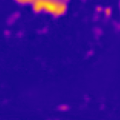}
    \includegraphics[width=0.15\columnwidth]{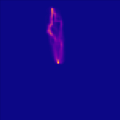}
    \includegraphics[width=0.15\columnwidth]{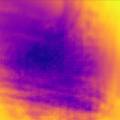}
    \includegraphics[width=0.15\columnwidth]{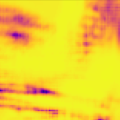}
    \includegraphics[width=0.15\columnwidth]{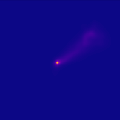}
    \includegraphics[width=0.01\columnwidth, height=0.14\columnwidth]{figs/revision/colorbar.png}
    \\
    \includegraphics[width=0.15\columnwidth]{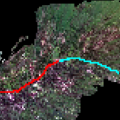}
    \includegraphics[width=0.15\columnwidth]{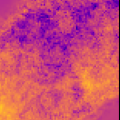}
    \includegraphics[width=0.15\columnwidth]{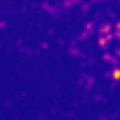}
    \includegraphics[width=0.15\columnwidth]{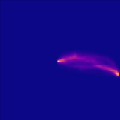}
    \includegraphics[width=0.15\columnwidth]{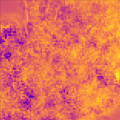}
    \includegraphics[width=0.15\columnwidth]{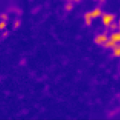}
    \includegraphics[width=0.15\columnwidth]{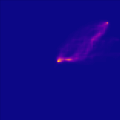}
    \includegraphics[width=0.15\columnwidth]{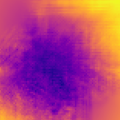}
    \includegraphics[width=0.15\columnwidth]{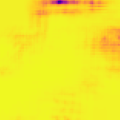}
    \includegraphics[width=0.15\columnwidth]{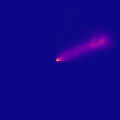}
    \includegraphics[width=0.01\columnwidth, height=0.14\columnwidth]{figs/revision/colorbar.png}
    \\
    \includegraphics[width=0.15\columnwidth]{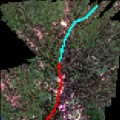}
    \includegraphics[width=0.15\columnwidth]{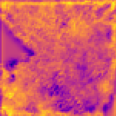}
    \includegraphics[width=0.15\columnwidth]{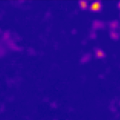}
    \includegraphics[width=0.15\columnwidth]{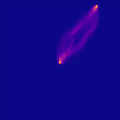}
    \includegraphics[width=0.15\columnwidth]{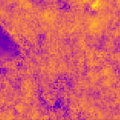}
    \includegraphics[width=0.15\columnwidth]{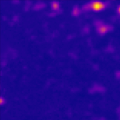}
    \includegraphics[width=0.15\columnwidth]{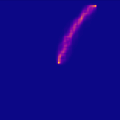}
    \includegraphics[width=0.15\columnwidth]{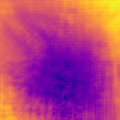}
    \includegraphics[width=0.15\columnwidth]{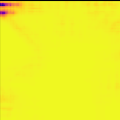}
    \includegraphics[width=0.15\columnwidth]{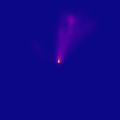}
    \includegraphics[width=0.01\columnwidth, height=0.14\columnwidth]{figs/revision/colorbar.png}
\\
    \begin{subfigure}{0.15\columnwidth}
        \includegraphics[width=1.0\columnwidth]{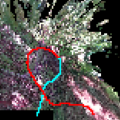}
        \caption{}
    \end{subfigure}
    \begin{subfigure}{0.15\columnwidth}
        \includegraphics[width=1.0\columnwidth]{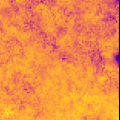}
        \caption{}
    \end{subfigure}
    \begin{subfigure}{0.15\columnwidth}
        \includegraphics[width=1.0\columnwidth]{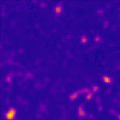}
        \caption{}
    \end{subfigure}
       \begin{subfigure}{0.15\columnwidth}
        \includegraphics[width=1.0\columnwidth]{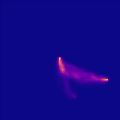}
        \caption{}
    \end{subfigure}
    \begin{subfigure}{0.15\columnwidth}
        \includegraphics[width=1.0\columnwidth]{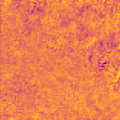}
        \caption{}
    \end{subfigure}
    \begin{subfigure}{0.15\columnwidth}
        \includegraphics[width=1.0\columnwidth]{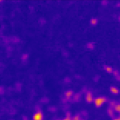}
        \caption{}
    \end{subfigure}
    \begin{subfigure}{0.15\columnwidth}
        \includegraphics[width=1.0\columnwidth]{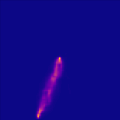}
        \caption{}
    \end{subfigure}
    \begin{subfigure}{0.15\columnwidth}
        \includegraphics[width=1.0\columnwidth]{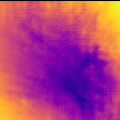}
        \caption{}
    \end{subfigure}
    \begin{subfigure}{0.15\columnwidth}
        \includegraphics[width=1.0\columnwidth]{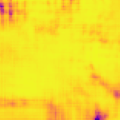}
        \caption{}
    \end{subfigure}
    \begin{subfigure}{0.15\columnwidth}
        \includegraphics[width=1.0\columnwidth]{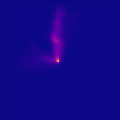}
        \caption{}
    \end{subfigure}
    \begin{subfigure}{0.01\columnwidth}
    \vspace{-5mm}
    \includegraphics[width=1.0\columnwidth, height=14\columnwidth]{figs/revision/colorbar.png}
    \end{subfigure}
\\
    \caption{Qualitative results on traversability cost learning and future trajectory prediction in Experiment 1. (a) are the input color maps overlaid with the demonstrated past (red) and future (cyan) trajectories. (b)-(d) are the qualitative results of our method with inertial feature learning: from (b) to (d) are respectively the inferred path reward maps, goal reward maps, and the SVF visualization of the path states representing the predicted future trajectory distributions. (e)-(g) are the corresponding results of our method without inertial feature learning, and (h)-(j) are of the baseline method with handcrafted kinematic features~\cite{zhang2018integrating}. The last row shows a failure case and the color map is shown in the right.
    }
    \label{fig:exp1-qualitative-results}
\end{figure*}

\subsection{Legged Robot Walking Dataset}

We collect data using a quadruped robot with a customized sensor suite operating in a campus environment including diverse scenes. The robot platform, an MIT Mini-Cheetah, and the data collecting environment are shown in Fig.~\ref{fig:data-collection}.
We manually teleoperate the robot to walk around by sending the commands of longitudinal and lateral velocity and yaw rate. 

We generate a dataset of over 1000 trajectories and their corresponding environmental features, IMU signals and AEC values using both exteroceptive and proprioceptive sensor measurements from different raw data sequences. For the environmental features, we generate the elevation map, elevation variance map and color map of a region of \mbox{8 m $\times$ 8 m} with 0.1 m resolution. Specifically, we use ORB-SLAM2~\cite{mur2017orb} to estimate the robot poses, and build a local robot-centric probabilistic elevation map using point clouds captured by the depth camera. Elevation mapping~\cite{fankhauser2018probabilistic} models the height measurements by a Gaussian distribution and fuses multi-frame measurements using a kalman filter. We generate the elevation and elevation variance map using the mean and variance of each grid's Gaussian distribution. The color map is generated by averaging the RGB values of all points falling into each grid.

The demonstrated trajectory is generated by projecting the robot poses onto the local grid map, including a past trajectory and a future trajectory. The past trajectory is a sequence of grid indices traversed by the robot during a period of time before map building, whereas the future trajectory is during the same period of time after map building. The IMU signals used for the inertial feature learning are collected within the past 0.5 $\sec$, and the AEC value is computed using the joint state measurements of 12 joints (3 joints per leg) along the past trajectory.
Examples of the generated data are visualized in Fig.~\ref{fig:network-input}. We use a 70/30 split to divide our data into training and test sets for the following experiments.

\subsection{Inertial Feature Learning Evaluation}

\begin{figure*}
    \centering
    \begin{minipage}{0.5\textwidth}
    \centering
    \includegraphics[width=0.14\columnwidth]{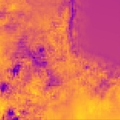} \hspace{-0.20cm}
    \includegraphics[width=0.14\columnwidth]{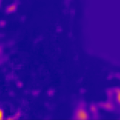} \hspace{-0.20cm}
    \includegraphics[width=0.14\columnwidth]{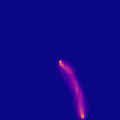} \hspace{-0.20cm}
    \includegraphics[width=0.14\columnwidth]{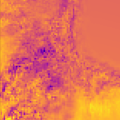} \hspace{-0.20cm}
    \includegraphics[width=0.14\columnwidth]{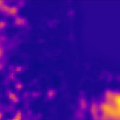} \hspace{-0.20cm}
    \includegraphics[width=0.14\columnwidth]{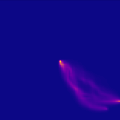} \hspace{-0.20cm}
    \includegraphics[width=0.14\columnwidth]{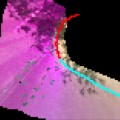} \\
    \includegraphics[width=0.14\columnwidth]{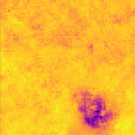} \hspace{-0.20cm}
    \includegraphics[width=0.14\columnwidth]{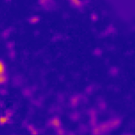} \hspace{-0.20cm}
    \includegraphics[width=0.14\columnwidth]{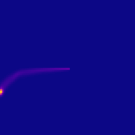} \hspace{-0.20cm}
    \includegraphics[width=0.14\columnwidth]{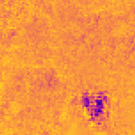} \hspace{-0.20cm}
    \includegraphics[width=0.14\columnwidth]{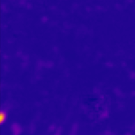} \hspace{-0.20cm}
    \includegraphics[width=0.14\columnwidth]{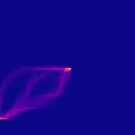} \hspace{-0.20cm}
    \includegraphics[width=0.14\columnwidth]{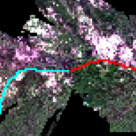}\\
    \includegraphics[width=0.14\columnwidth]{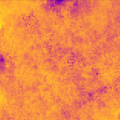} \hspace{-0.20cm}
    \includegraphics[width=0.14\columnwidth]{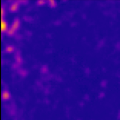} \hspace{-0.20cm}
    \includegraphics[width=0.14\columnwidth]{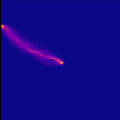} \hspace{-0.20cm}
    \includegraphics[width=0.14\columnwidth]{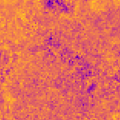} \hspace{-0.20cm}
    \includegraphics[width=0.14\columnwidth]{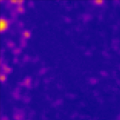} \hspace{-0.20cm}
    \includegraphics[width=0.14\columnwidth]{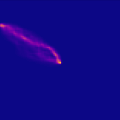} \hspace{-0.20cm}
    \includegraphics[width=0.14\columnwidth]{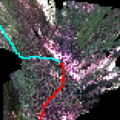}
    \\
    \includegraphics[width=0.14\columnwidth]{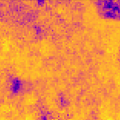} \hspace{-0.20cm}
    \includegraphics[width=0.14\columnwidth]{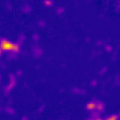} \hspace{-0.20cm}
    \includegraphics[width=0.14\columnwidth]{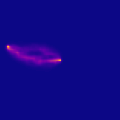} \hspace{-0.20cm}
    \includegraphics[width=0.14\columnwidth]{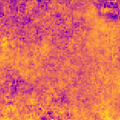} \hspace{-0.20cm}
    \includegraphics[width=0.14\columnwidth]{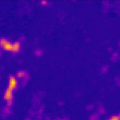} \hspace{-0.20cm}
    \includegraphics[width=0.14\columnwidth]{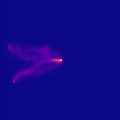} \hspace{-0.20cm}
    \includegraphics[width=0.14\columnwidth]{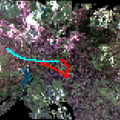}
    \end{minipage}
    \begin{minipage}{0.48\textwidth}
    \centering
    \includegraphics[width=0.75\columnwidth]{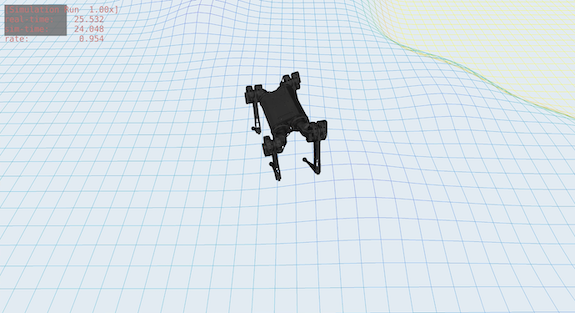} \\
    \vspace{0.07cm}
    \includegraphics[width=0.24\columnwidth, trim={0cm 0cm 0cm 0.3cm}, clip]{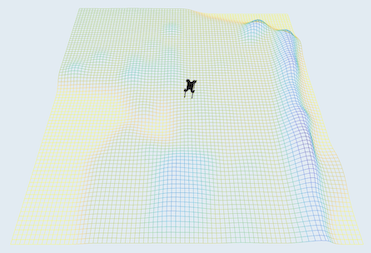}
    \includegraphics[width=0.24\columnwidth]{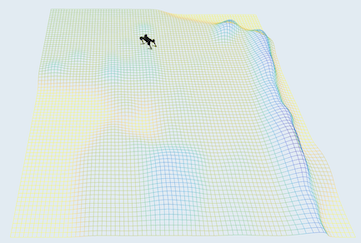} 
    \includegraphics[width=0.24\columnwidth, trim={0cm 0cm 1cm 0cm}, clip]{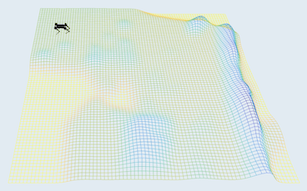}
    \end{minipage}
    \caption{Left block: Qualitative results on traversability cost learning and future trajectory prediction in Experiment 2. From left to right, the first three columns are the inferred path reward maps, goal reward maps and the predicted future trajectory distributions obtained using the MEDIRL algorithm. The next three columns are those obtained using our proposed T-MEDIRL algorithm with energy-based trajectory ranking. The last column is the corresponding input color maps overlaid with demonstrated past (red) and future (cyan) trajectories. The magenta region in the color maps is due to a hardware issue of the camera. Right block: Mini-Cheetah Simulator. The bottom row shows the robot is following the optimal future trajectory.}
    \label{fig:exp2-qualitative-results}
\end{figure*}

This experiment aims to evaluate the performance of the proposed reward network in reward learning, i.e., how well the learned reward function explains the expert demonstration. Therefore, we employ two commonly used metrics in IRL problems for quantitative evaluation: the Negative Log-Likelihood (NLL) of the expert demonstration under the learned policy, and the Hausdorff Distance (HD)~\cite{kitani2012activity} between demonstrations and trajectories sampled with the learned policy. Here, the learned policy is obtained via value iteration given the learned reward function. For a fair comparison, all networks in this experiment are trained using the MEDIRL framework under the same MDP formulation.

We first conduct an ablation study on the environmental branch to ensure the performance of terrain feature extraction. We compare three different architectures used in literatures: ResNet~\cite{deo2020trajectory}, UNet~\cite{ronneberger2015u}, and ResUNet~\cite{zhang2018road}. The quantitative results of reward learning using only environmental features are given in the upper part of Table~\ref{table1}. In the rest of the experiment, we use the best performing ResUNet as our environmental branch. Next, we compare the performance of our method with the inertial feature learning and of a baseline approach using handcrafted kinematic features~\cite{zhang2018integrating}. For additional comparison with the networks without inertial feature learning, we list the quantitative results in the lower part of Table~\ref{table1}. The superiority of our method for legged robots can be seen in Table~\ref{table1}, where our network with the inertial branch has the best prediction results using both metrics.

The qualitative results are given in Fig.~\ref{fig:exp1-qualitative-results}. The first column are the input color maps for each scene, overlaid with past trajectories (red) and ground truth future trajectories (cyan). The results of each method are then shown side by side. The color of each pixel in the path reward map represents the reward of traversing that state (the negative path reward is the traversability cost of that state), whereas the goal reward map indicates the reward of terminating at that state. The corresponding path SVF visualization indicates the distributions of the predicted future trajectories under the inferred rewards.
The qualitative results also show our method with inertial feature learning generally has the best performance in reward learning and trajectory prediction compared with the baseline and our method without inertial features, although fails in certain scenarios (last row of Fig.~\ref{fig:exp1-qualitative-results}).

\begin{table}[t]
\caption{Traversability Cost Learning/Future Trajectory Prediction Performance Quantitative Comparison in Experiment 1. For environmental branch ablation study, only environmental features are used as the network input. By learning and incorporating robot inertial features in the proposed reward network, the inferred reward function (traversability cost) becomes robot-state-dependent, thus achieving better performance than purely environment-dependent reward.}
\label{table1}
\centering
{\scriptsize
\begin{tabular}{l|l|c|c}
\toprule
& Method & NLL $\downarrow$ & HD $\downarrow$ \\
\hline
\multirow{3}{*}{Env. Branch Ablation} & \rd{ResNet~\cite{deo2020trajectory}} & \rd{0.9490} & \rd{13.4957} \\
& \rd{UNet~\cite{ronneberger2015u}} & \rd{0.9016} & \rd{12.0014} \\
& \rd{ResUNet~\cite{zhang2018road}} & \rd{\bf 0.8419} & \rd{\bf 9.8219} \\
\hline
\multirow{2}{*}{\rd{Inertial Feat. Evaluation}} & \rd{Dilated CNN w/ kinematics~\cite{zhang2018integrating}} & \rd{0.8821} & \rd{10.1953} \\
& \rd{ResUNet w/ inertial features} & \rd{\bf 0.7915} & \rd{\bf 8.4012} \\
\bottomrule
\end{tabular}
}
\end{table}

\subsection{Energy-based Reward Extrapolating Evaluation}

We evaluate the effect of locomotion energy ranked reward extrapolation in this experiment. We use the AEC value to rank the demonstrated trajectories in the training set, and train the proposed reward network using the proposed T-MEDIRL framework. For comparison, we choose the same reward network trained using the MEDIRL framework as the baseline. The qualitative results are given in the left block of Fig.~\ref{fig:exp2-qualitative-results}. We notice that the path reward maps learned by T-MEDIRL are less smooth and more likely to induce a multimodal distribution over path and goal states. Given the difference in performance, the less smooth reward map can be associated with their only difference, namely the trajectory ranking loss. To further evaluate the performance, we quantitatively evaluate the predicted trajectories by both methods.

As the purpose of introducing energy-based trajectory ranking loss is to learn a reward/policy better than demonstration, quantitative metrics other than NLL and HD are needed in this experiment. In classical IRL problems, the quality of the predicted reward can be evaluated by comparing with the ground truth reward. Due to the lack of ground truth reward in real-world problems, we introduce the following two metrics for evaluation.

The first one is classification accuracy. 
The reward learning in T-MEDIRL framework can be formulated as a classification problem, where the trajectory with higher predicted return is classified as higher ranked given the predicted reward. Thus, we compare the classification accuracy on the test set for both methods. Specifically, for any pair of test trajectories, we compute their returns from the predicted reward maps, if the one having higher return corresponds to a smaller AEC value, it is considered to be correct. The accuracy is computed by dividing the total number of correct classifications by the total number of comparisons.

Secondly, to evaluate whether the reward map learned by our T-MEDIRL algorithm can lead to a more energy-efficient policy, we generate optimal future trajectories for both methods, and compare the AEC values of the planned trajectories. To this end, we use a simulation environment for Mini-Cheetah that provides the robot model, a state estimator and a controller. We implement a simple PID-based waypoint follower to send motion commands to the controller. For each planned trajectory of the test data, we load the corresponding elevation map to the simulator to simulate the terrain conditions, and make the robot follow the planned trajectory under the trot mode (right block of Fig.~\ref{fig:exp2-qualitative-results}). The yaw commands are automatically generated to follow the robot's moving direction.

\begin{table}[t]
\caption{Traversability Cost Learning/Future Trajectory Prediction Performance Quantitative Comparison in Experiment 2. We log the simulated robot joint states during the trajectory following for all planned trajectories in the test set, using a sampling time of $2~\mathrm{ms}$. For each method, the AEC computed for all planned trajectories is reported. Using a $24 \mathrm{V}$, $6 \mathrm{Ah}$ battery, the difference shown below in AEC corresponds to about seven minutes extra operation. }
\label{table2}
\centering
{\scriptsize
\begin{tabular}{l|c|c}
\toprule
Method & ours w/ MEDIRL~\cite{wulfmeier2017large} & ours w/ T-MEDIRL \\
\hline
NLL $\downarrow$ & \rd{\bf 0.7734} & \rd{0.8132} \\
\hline
HD $\downarrow$ & \rd{\bf 8.1460} & \rd{9.9126} \\
\hline
Accuracy $\uparrow$ & \rd{40.01\%} & \bf \rd{64.12\%} \\
\hline
AEC $\downarrow$ & \rd{$4.634\times 10^{-2} \mathrm{J}$} & \rd{\bf $\bf 4.179\times 10^{-2} \mathrm{J}$} \\
\bottomrule
\end{tabular}
}
\end{table}

The quantitative results are given in Table~\ref{table2}, where we report the classification accuracy and AEC besides NLL and HD. As expected, MEDIRL has better prediction performance (shown by its lower NLL and HD) compared with T-MEDIRL, indicating that MEDIRL learns a reward function that better explains the demonstration, and a policy closer to the demonstrated behavior. In contrast, the reward maps learned using T-MEDIRL lead to a much higher classification accuracy, which shows the effect of the energy-based trajectory ranking loss in reward regulation. Last but not least, the lower simulated AEC of the optimal trajectories planned by T-MEDIRL directly shows that the reward maps learned by T-MEDIRL are extrapolated beyond the suboptimal demonstrations and lead to a more energy-efficient policy than demonstration. 

\subsection{Discussion}
The limitation of this work is that the simplified discrete states and actions are unable to fully capture the agile motion capability of a legged robot. However, adding more dimensions to the state, such as orientation, velocities, and higher-order derivatives, comes with an exponential increase in computation complexity. To model complex legged robot locomotion as a tractable IRL problem, designing a hierarchical action and state space, e.g.~\cite{kolter2007hierarchical}, is an interesting future research direction.

The following framework is a feasible research direction to deploy this work for online exploration missions. The IIG algorithm in~\cite{ghaffari2019sampling} does not require any goals to compute exploration policies. Combined with a Model Predictive Control (MPC)~\cite{teng2021toward,teng2022error}, it can provide an integrated kinodynamic planner that takes the robot stability, control constraints, and the value of information from sensory data into account. In particular, the developed work here can serve as the data-driven information function in the IIG algorithm to compute a partial ordering of future states.

\section{Conclusion}
\label{sec:conclusion}

This work tackles the problem of terrain traversability modeling for legged robots using deep IRL. Two issues have been considered in this work. First, legged robots with high mobility tend to have more sudden motions. Existing DIRL works using purely environmental features or handcrafted kinematic features might fail to capture these movements. The proposed deep reward network can effectively learn robot inertial features from proprioceptive sensory data and integrate the learned features into the reward map. 
Secondly, we have proposed a T-MEDIRL algorithm and trajectory ranking strategy using energy to deal with the suboptimality of legged robot demonstrations. By introducing a preference label for each demonstrated trajectory, the inferred reward map can lead to a more energy-efficient policy.

{\scriptsize
\balance
\bibliographystyle{bib/IEEEtranN}
\bibliography{bib/IEEEabrv,bib/strings-abrv,bib/ieee-abrv,bib/refs}
}

\end{document}